\def\year{2022}\relax
\documentclass[letterpaper]{article} 
\usepackage{aaai22}  
\usepackage{times}  
\usepackage{helvet}  
\usepackage{courier}  
\usepackage[hyphens]{url}  
\usepackage{graphicx} 
\usepackage{natbib}  
\usepackage{caption} 
\DeclareCaptionStyle{ruled}{labelfont=normalfont,labelsep=colon,strut=off} 
\usepackage{algorithm}
\usepackage{algorithmic}

\usepackage{newfloat}
\usepackage{listings}
\floatstyle{ruled}
\newfloat{listing}{tb}{lst}{}
\floatname{listing}{Listing}
\title{Domain Adaptive Pretraining for Multilingual Acronym Extraction}
\author{
    Usama Yaseen, \textsuperscript{\rm 1,2}  
    Stefan Langer \textsuperscript{\rm 1,2}
}
\affiliations{
    \textsuperscript{\rm 1} Technology, Siemens AG Munich, Germany\\
    \textsuperscript{\rm 2} CIS, University of Munich (LMU) Munich, Germany\\
    usama.yaseen@siemens.com, langer.stefan@siemens.com
}

\begin{document}

\maketitle

\begin{abstract}

This paper presents our findings from participating in the multilingual acronym extraction shared task SDU@AAAI-22. The task consists of acronym extraction from documents in $6$ languages within scientific and legal domains. To address multilingual acronym extraction we employed BiLSTM-CRF with multilingual XLM-RoBERTa embeddings. We pretrained the XLM-RoBERTa model on the shared task corpus to further adapt XLM-RoBERTa embeddings to the shared task domain(s). Our system (team: SMR-NLP) achieved competitive performance for acronym extraction across all the languages.

\end{abstract}

\section{Introduction}

\begin{figure}[t]
\centering
    \includegraphics[scale=0.50]{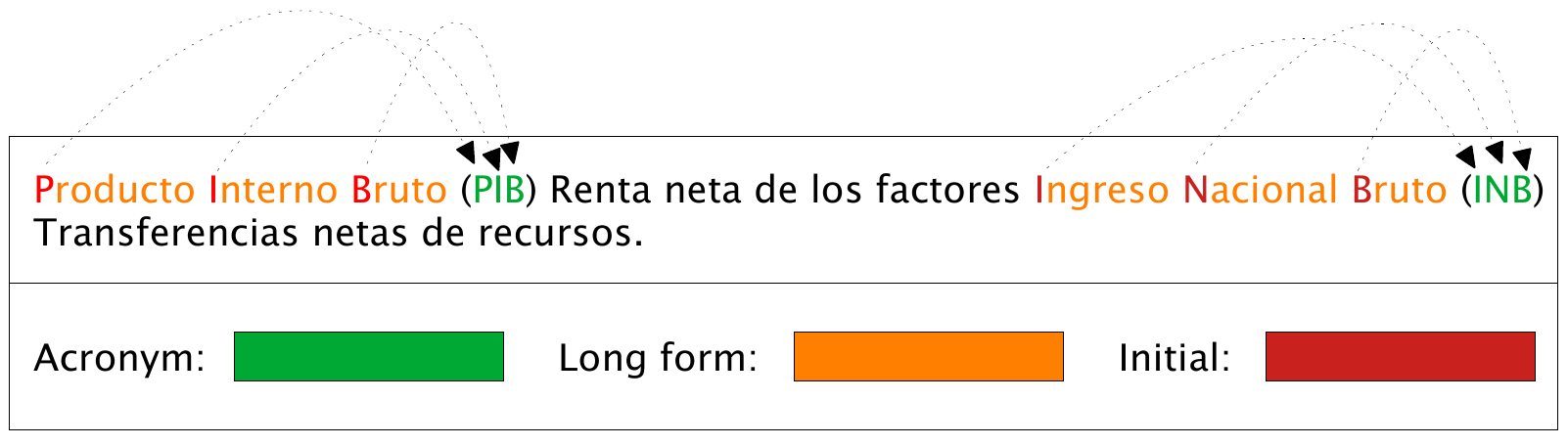}
    \caption{ An example from the Spanish acronym extraction dataset. In the figure, the Green text represents acronyms, orange text represents long term, and red text represents initials. Also, the black lines indicate the correspondence between initials and acronyms.}
    \label{fig:ae-example}
\end{figure}

The number of scientific papers published every year is growing at an increasing rate \cite{BornmannM15}. The authors of the scientific publications employ abbreviations as a tool to make technical terms less verbose. The abbreviations take the form of acronyms or initialisms. We refer to the abbreviated term as ``acronym" and we refer to the full term as the ``long form" (see Figure \ref{fig:ae-example}). On one hand, the acronyms enable avoiding frequently used long phrases making writing convenient for researchers but on the other hand they pose a challenge to non-expert human readers. This challenge is heightened by the fact that the acronyms are not always standard written, e.g. XGBoost is an acronym of eXtreme Gradient Boosting \cite{ChenG16}. Following the increase of scientific publications, the number of acronyms is enormously increasing as well \cite{Barnett2020TheGO}. Thus, automatic identification of acronyms and their corresponding long forms is crucial for scientific document understanding tasks.

The existing work in acronym extraction consists of carefully crafted rule-based methods \cite{SchwartzH03, OkazakiA06} and feature-based methods \cite{KuoLLH09, LiuLH17}. These methods typically achieve high precision as they are designed to find long form, however, they suffer from low recall \cite{HarrisS19}. Recently, Deep Learning based sequence models like LSTM-CRF \cite{VeysehDTN20} have been explored for the task of acronym extraction, however, these methods require large training data to achieve optimal performance. One of the major limitations of existing work in acronym extraction is that most prior work only focuses on the English language.

\section{Task Description and Contributions}

We participate in the {Acronym Extraction} task \cite{veyseh-et-al-2022-Multilingual} organized by the Scientific Document Understanding workshop 2022 (SDU@AAAI-22). The task consists of identifying acronyms (short-forms) and their meanings (long-forms) from the documents in six languages including Danish (da), English (en), French (fr), Spanish (es), Persian (fa) and Vietnamese (vi). The task corpus \cite{veyseh-et-al-2022-MACRONYM} consists of documents from the scientific (en, fa, vi) and legal domains (da, en, fr, es).

Following are our multi-fold contributions:

1. We model multilingual acronym extraction as a sequence labelling task and employed contextualized multilingual {\it XLM-RoBERTa} embeddings \cite{ConneauKGCWGGOZ20}. Our system consists of a single model for multilingual acronym extraction and hence is practical for real-world usage.

2. We investigated domain adaptive pretraining of {\it XLM-RoBERTa} on the task corpus, which resulted in improved performance across all the languages.

\section{Methodology}

\begin{figure*}[t]
    \centering
    \includegraphics[scale=0.74]{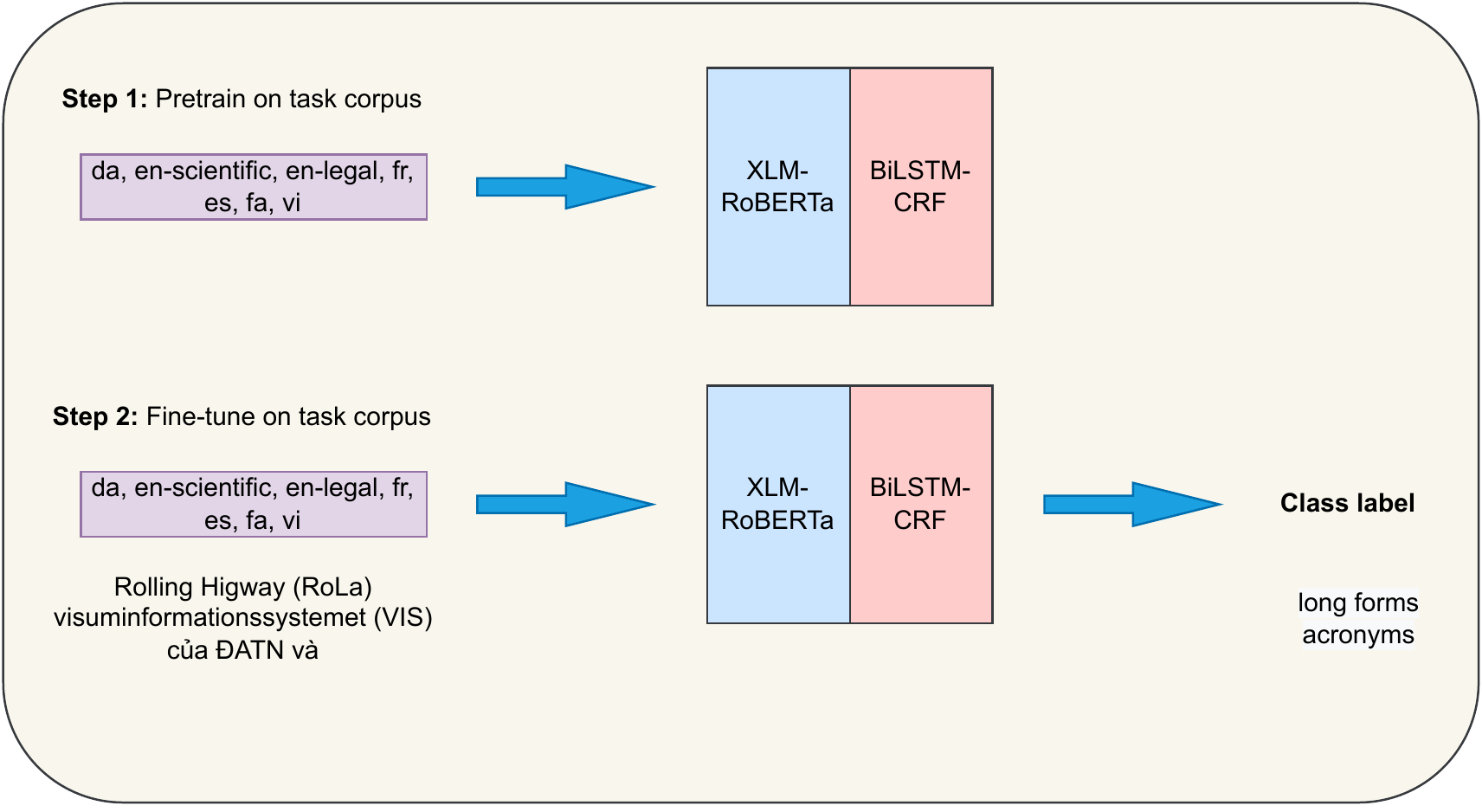}
    \caption{An illustration of domain adaptation for multilingual acronym extraction model.}
    \label{fig:ae-architecture}
\end{figure*}

\begin{table*}[t]
\centering
\resizebox{.95\textwidth}{!}{
\begin{tabular}{| c | c | c | c | c | c | c | c | c | c | c}
\hline
& {\bf epochs} & \textbf{all} & \textbf{da} & \textbf{en-sci} & \textbf{en-leg} & \textbf{fr} & \textbf{fa} & \textbf{es} & \textbf{vi} \\
& & P/R/F1 & P/R/F1 & P/R/F1& P/R/F1 & P/R/F1 & P/R/F1 & P/R/F1 & P/R/F1\\ \hline
\multicolumn{10}{c}{{\it dev}}  \\ \hline
r1 & 0 & .841/.868/.854 & .825/.833/.829 & .727/.750/.738 & .758/.784/.771 & .738/.742/.740 & .619/.539/.576 & .820/.871/.845 & .375/.547/.445 \\
r2 & 1 & .855/.876/.866 & .826/.833/.830 & .747/.757/.752 & .786/.793/.789 & .756/.750/.753 & .644/.560/.599 & .832/.872/.852 & .385/.615/.474 \\
r3 & 3 & {\bf .857/.878/.868} & {\bf .827/.833/.830} & {\bf .750/.759/.755} & {\bf .789/.795/.792} & {\bf .788/.751/.754} & {\bf .665/.557/.606} & {\bf .832/.873/.852} & {\bf .408/.689/.512} \\
r4 & 3 & -  & .77/.773/.775 & .617/.703/.650 & .677/.677/.677 & .715/.733/.724 & .864/.294/.439 & .823/.850/.836 & .623/.074/.132 \\ \hline
\multicolumn{10}{c}{{\it test}}  \\ \hline
r5 & 3 & - & .825/.833/.829 & .727/.750/.738 & .758/.784/.771 & .738/.742/.740 & .619/.539/.576 & .820/.871/.845 & .375/.547/.445 \\ \hline

\end{tabular}
}
\caption{F1-score on the development set (r1-r4) and test set (r5). Here, {\it epochs}: number of pretraining epochs for XLM-RoBERTa on the task corpus, {\it eng-sci}: english scientific domain, {\it eng-leg}: english legal domain, {\it all}: all languages combined.}
\label{table:results}
\end{table*}

In the following sections we discuss our proposed model for acronym extraction.

\subsection{Multilingual Acronym Extraction}

Our sequence labelling model follows the well-known architecture \cite{LampleBSKD16} with a bidirectional long short-term memory (BiLSTM) network and conditional random field (CRF) output layer \cite{LaffertyMP01}. In order to address the multilingual aspect of the task we employed contextualized multilingual {\it XLM-RoBERTa} embeddings \cite{ConneauKGCWGGOZ20} in all the experiments.

\subsection{Domain Adaptive Pretraining}

The original  {\it XLM-RoBERTa} embeddings \cite{ConneauKGCWGGOZ20} are trained on the filtered CommonCrawl data (General domain), whereas the data of the shared task comprises documents from scientific and legal domains. In order to better adapt the contextualized representation to the target scientific and legal domain, we further pretrained the original XLM-RoBERTa model on the corpus data (see Figure \ref{fig:ae-architecture}). Our experiments demonstrate improved performance on the task of acronym extraction due to the domain adaptive pretraining across all the languages.

\section{Experiments and Results}

\subsection{Dataset}

\begin{table}[t]
\centering
\resizebox{.30\textwidth}{!}{
\begin{tabular}{l | c | c | c}
{\bf language} & \textbf{train} & \textbf{dev} & \textbf{test} \\
da & 3082 & 385 & 386 \\
eng-scientific & 3980 & 497 & 498 \\
eng-legal & 3564 & 445 & 446 \\
fr & 7783 & 973 & 973 \\
es & 5928 & 741 & 741 \\
fa & 1336 & 167 & 168 \\
vi & 1274 & 159 & 160 \\
\end{tabular}
}
\caption{Sentence counts of train and development set across the languages.}
\label{table:dataset}
\end{table}

\begin{table}[t]
\centering
\resizebox{.30\textwidth}{!}{
\begin{tabular}{l | l}
{\bf Hyperparameter} & \textbf{Value}\\
hidden size & $256$\\
learning rate & $5.0e-6$\\
training epochs & $20$\\
pretraining epochs & $3$\\
\end{tabular}
}
\caption{Hyperparameter settings for acronym extraction.}
\label{table:hyperparameters}
\end{table}

Table \ref{table:dataset} reports sentence counts in the train and development set for all the languages. Persian and Vietnamese have substantially low sentences compared to the rest of the languages in the corpus. As a pre-processing step, we used {\it spaCy} \cite{matthew_honnibal_spacy} to perform word tokenization and POS tagging.

We do not apply any strategy to explicitly account for low training data of Persian and Vietnamese. Table \ref{table:hyperparameters} lists the best configuration of hyperparameters. We compute macro-averaged F1-score using the script provided by the organizers on the development set \footnote{https://github.com/amirveyseh/AAAI-22-SDU-shared-task-1-AE/blob/main/scorer.py}. We employ early stopping and report the F1-score on the test set using the best performant model on the development set.

\subsection{Results}

Table \ref{table:results} reports the F1-score on the development and test set for all the languages. As a baseline experiment, we combined the training data for all the languages and trained a BiLSTM-CRF model using the pretrained multilingual XLM-Roberta\footnote{\url{https://huggingface.co/xlm-roberta-base}} embeddings (row r1). This achieves the overall F1-score of $0.854$.

We pretrained XLM-Roberta model for $1$ epoch on the task corpus using train and development set, which results in $0.1$ points improvement in the overall F1-score leading to the F1-score of $0.866$ (row r2). Increasing the pretraining epochs to $3$ results in an improvement of additional $0.1$ points in the overall F1-score (row r3).

We also experimented with training the individual models for each language (including separate models for English scientific and English legal). This results in a significant decrease in F1-score for all the languages (on average $0.12$ points in F1-score, see row r4). This demonstrates that BiLSTM-CRF with multilingual XLM-Roberta embeddings performs best when trained with several languages together enabling effective cross-lingual transfer. 

The F1-score of our submission on the test set are reported in row r5. Our test submission achieves the F1-score similar to the development set for all the languages demonstrating effective generalization on the test set; Vietnamese is an exception where F1-score on the test set is significantly worse than the F1-score on the development set (see rows r5 vs r3).

\section{Conclusion}

In this paper, we described our system with which we participate in the multilingual acronym extraction shared task organized by the Scientific Document Understanding workshop 2022 (SDU@AAAI-22). We formulate multlilignual acronym extraction in $6$ languages and $2$ domains as a sequence labelling task and employed BiLSTM-CRF model with multilingual XLM-RoBERTa embeddings. We pretrained XLM-RoBERTa model on the target scientific and legal domain to better adapt multilingual XLM-RoBERTa embeddings for the target task. Our system demonstrates competitive performance on the multilingual acronym extraction task for all the languages. In future, we would like to improve error analysis to further enhance our multilingual acronym extraction models.

\section{Acknowledgments}

This research was supported by the Federal Ministry for Economic Affairs and Energy ( Bundesministerium für Wirtschaft und Energie: \url{https://bmwi.de}), grant 01MD19003E (PLASS: \url{https://plass.io}) at Siemens AG (Technology), Munich Germany.

\bibliography{aaai22.bib}
\end{document}